\documentclass[letterpaper, 10 pt, conference]{ieeeconf}  

\IEEEoverridecommandlockouts                              

\usepackage{bm}
\usepackage{algorithmic}
\usepackage{algorithm}

\usepackage{amsmath,amssymb,amsfonts}

\usepackage{graphicx}
\graphicspath{ {./imgs/} }
\usepackage[caption=false]{subfig}

\usepackage{booktabs}

\usepackage{flushend}
\usepackage{cite}
\makeatletter
\let\NAT@parse\undefined
\makeatother
\usepackage[colorlinks,linkcolor=black,anchorcolor=black,citecolor=black,urlcolor=black,hyperfootnotes=true]{hyperref}   
\usepackage[all]{hypcap} 

\title{\LARGE \bf
Indoor Localization for Quadrotors using Invisible Projected Tags
}

\author{Jinjie Li$^{1}$, Liang Han$^{2*}$, Zhang Ren$^{1}$
\thanks{$^{1}$J. Li and Z. Ren are with the School of Automation Science and Electrical Engineering, Beihang University, Beijing, 100191, China 
{\tt\small lijinjie@buaa.edu.cn};
{\tt\small renzhang@buaa.edu.cn}}

\thanks{$^{2}$L. Han is with the Sino-French Engineer School, Beihang University, Beijing, 100191, China. 
{\tt\small liang\_han@buaa.edu.cn}}
}

\begin{document}

\maketitle
\thispagestyle{empty}
\pagestyle{empty}

\begin{abstract}
    Augmented reality (AR) technology has been introduced into the robotics field to narrow the visual gap between indoor and outdoor environments. However, without signals from satellite navigation systems, flight experiments in these indoor AR scenarios need other accurate localization approaches. This work proposes a real-time centimeter-level indoor localization method based on psycho-visually invisible projected tags (IPT), requiring a projector as the sender and quadrotors with high-speed cameras as the receiver. The method includes a modulation process for the sender, as well as demodulation and pose estimation steps for the receiver, where screen-camera communication technology is applied to hide fiducial tags using human vision property. Experiments have demonstrated that IPT can achieve accuracy within ten centimeters and a speed of about ten FPS. Compared with other localization methods for AR robotics platforms, IPT is affordable by using only a projector and high-speed cameras as hardware consumption and convenient by omitting a coordinate alignment step. To the authors' best knowledge, this is the first time screen-camera communication is utilized for AR robot localization.

\end{abstract}

\bigskip


\section{Introduction}

The deployment of small and intelligent aerial robots has significantly influenced civilian tasks such as transportation, environment preservation, and agriculture \cite{floreano_science_2015}. These complex tasks present scientific and technical challenges on perception, decision making, and agile flight control.
For one of the most widely used aerial robots, the quadrotor, many researchers carry out flight tests indoors to reduce the experimental risk. However, two problems exist: first, indoor environments are just simple abstractions of actual mission scenarios, unable to provide enough visual information; second, quadrotors require other methods to localize without global navigation satellite system signals.



For the first problem, researchers have combined augmented reality (AR) technology with mobile robot research in recent years \cite{makhataeva_augmented_2020}. Reina et al. \cite{reina_augmented_2015} proposed the concept ``virtual sensors'' and presented an AR-based sensing framework to control the swarm e-puck robots.
Omidshafiei et al. \cite{omidshafiei_measurable_2016} designed a novel robotics platform for hardware-in-the-loop experiments, called measurable augmented reality for prototyping cyber-physical systems (MAR-CPS). The MAR-CPS supplied visual information by projecting a mission scenario and robot states onto the ground through several projectors.
However, the above AR robotics platforms all need additional devices to solve the second problem, indoor localization. Specifically, common solutions such as motion capture system or Ultra-Wide Band (UWB) devices are used \cite{foehn_time-optimal_2021}, \cite{shule_uwb-based_2020}. Nevertheless, the former is high-priced, and the latter cannot measure orientation and is easily affected by metal surfaces \cite{wang_research_2018}. Furthermore, separating the localization system and the displaying system adds an extra step to align two coordinate systems \cite{omidshafiei_measurable_2016}. This step causes both hardware consumption and operation complexity. Therefore, the AR robotics platforms require a low-cost and easy-to-use indoor localization method for wide use.

\setlength{\textfloatsep}{8pt plus 1.0pt minus 2.0pt}
\begin{figure}[t]
    \centerline{\includegraphics[scale=0.24]{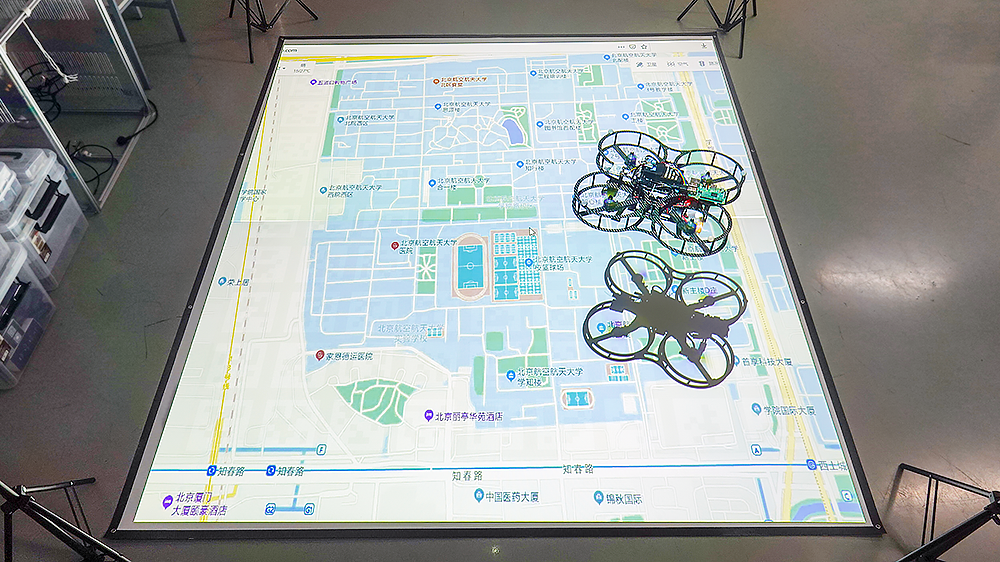}} 
    \caption{A quadrotor is flying on a projected AR scenario, using IPT to collect position data. Requiring no additional equipment, the quadrotor obtains pose from only projected images through a downward camera.}
    \label{fig:real_situation}
\end{figure}

In this work, we propose a localization method based on invisible projected tags (IPT), under the idea to merge the visualization and localization processes.
The main difficulty of IPT is to embed position information into projected images without influencing the visual effect. Thus, screen-camera communication is utilized to hide fiducial tags in the high-frequency flashing frames. Finally, experiments (Fig. \ref{fig:real_situation}) demonstrate that IPT method can provide centimeter-level position but requires just projectors and cameras as the hardware cost.



The main contributions of this article are:
\begin{itemize}
    \item IPT, a real-time centimeter-level localization method based on the invisible projected fiducial tags,
    \item an IPT-based indoor AR flight platform for quadrotors,
    \item experiments to verify feasibility and test performance.
\end{itemize}






\section{Related Work}

Indoor localization methods for quadrotors can be divided into non-cooperative localization technology and cooperative localization technology. For non-cooperative localization technology such as simultaneous localization and mapping (SLAM), the requirements of heavy sensors or complex algorithms make it unsuitable for quick-start indoor flights. For cooperative localization technology, the extensively used methods for indoor flight include motion capture system, base station system, and 2D fiducial markers \cite{preiss_crazyswarm_2017,yu_novel_2019,beul_fast_2018,xing_marker-based_2018}.


The motion capture system can measure position and orientation at the submillimeter level with high reliability, but the high price limits it in only well-funded research institutions or the film industry. For the base station system, the popular Ultra-Wide Band (UWB) technology can achieve positioning accuracy within ten centimeters at a low cost \cite{zafari_survey_2019}, but the lack of orientation measurement and the sensitivity to metal surfaces limit its use \cite{wang_research_2018}. The fiducial marker system achieves centimeter-level localization accuracy robustly at the cost of just a camera and several printed papers, making it the choice of many robotic researchers \cite{serra_landing_2016}, \cite{wang_acmarker_2020}. There has been sufficient research on the design, detection, and decoding processes of visual fiducial markers, such as ARTag \cite{garrido-jurado_automatic_2014}, ArUco 3 \cite{romero-ramirez_speeded_2018}, and AprilTag 3 \cite{krogius_flexible_2019}. For tag-based pose estimation problem, also named as \textit{Solve Pose from $n$ Points} (SolvePnP) problem, many algorithms have been proposed and integrated into the OpenCV library, such as EPnP \cite{lepetit_epnp_2009}, IPPE \cite{collins_infinitesimal_2014}, and SQPnP \cite{vedaldi_consistently_2020}.
The motivation is to design an inexpensive localization method for indoor AR robotics systems, and hence the tag-based approach is chosen in this work.

In order to hide the fiducial tags in the projected video, another technique used is screen-camera communication, which conveys invisible information using the property of human eyes.
In this field, the primary metrics include invisibility, throughput, and reliability \cite{zhang_chromacode_2021}. For marker-based localization, the information transmitted is relatively limited, and hence the algorithms with excellent invisibility and reliability are focused, including HiLight \cite{li_real-time_2015}, TextureCode \cite{nguyen_high-rate_2016}, and ChromaCode \cite{zhang_chromacode_2021}. HiLight hides information in the time domain through rapid brightness changes. TextureCode goes deeper by hiding data in the time and spatial domain, based on the principle that human eyes have different sensitivities to image areas with various texture densities. ChromaCode achieves better invisibility by converting the image to the CIELAB color space for lightness change, matching the human eye's sensitivity to different colors.
The advantages of these algorithms can be taken to design IPT method.


\section{System Overview}

\begin{figure}[t]
    \centerline{\includegraphics[scale=0.49]{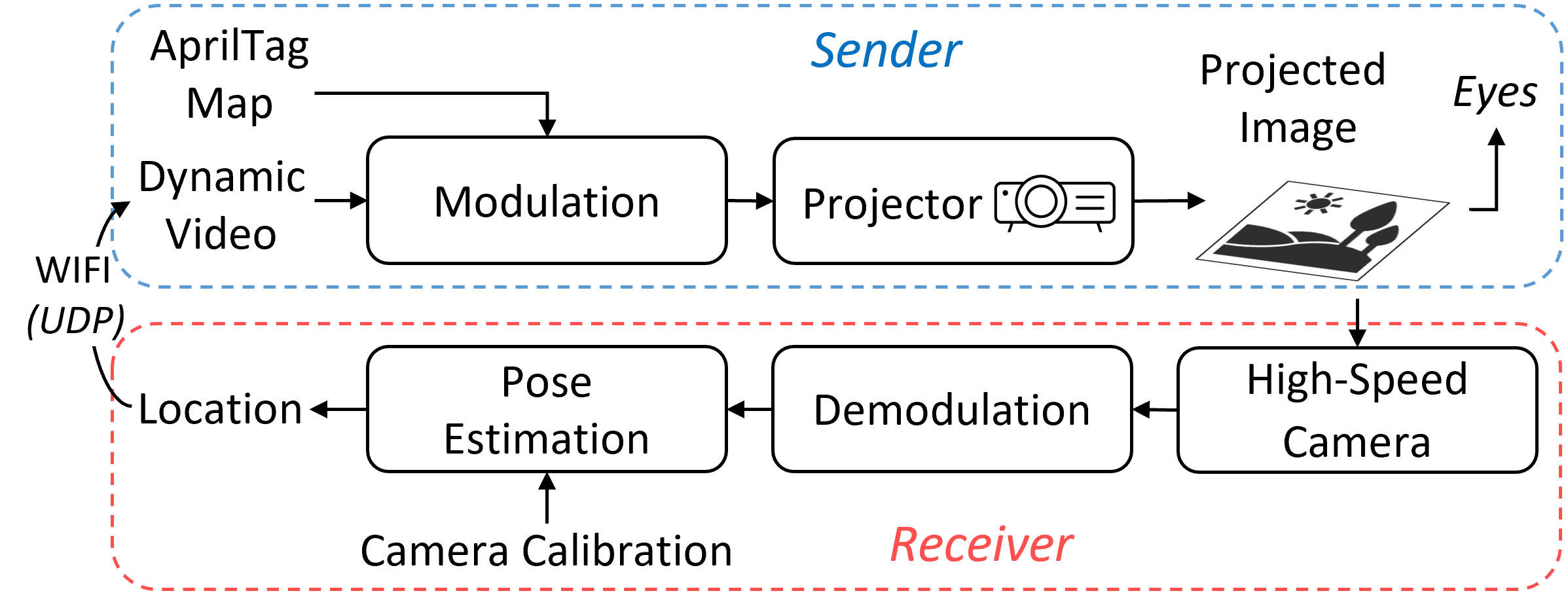}}
    \caption{The overall workflow of IPT. A video is first embedded with a bundle of invisible AprilTags. The sender side then projects a video onto the ground. Next, the receiver demodulates the tags through the frames from a high-speed camera and obtains its location. Finally, the location is sent to a computer to generate an AR dynamic scenario.}
    \label{fig:workflow}
\end{figure}

This work designs an indoor AR flight platform to test the performance of IPT method, and its workflow is illustrated in Fig. \ref{fig:workflow}. The platform is divided into a \textit{Sender} part and a \textit{Receiver} part. On the sender side, first, the \textit{Modulation} algorithm embeds a visual fiducial tag map into a dynamic video; second, a projector projects the modulated video to the ground, providing a visually complex AR environment for both quadrotors and the audience.
On the receiver side, a high-speed camera is used to capture the image information. The \textit{Demodulation} algorithm is then executed to obtain tag corners' three-Dimensional (3D) world coordinates and their two-Dimensional (2D) projections from the captured video. Finally, these coordinates are input to the \textit{Pose Estimation} step with camera parameters to compute pose. Through WIFI (UDP), the pose data are transmitted to the sender to change the video content for AR effects. Viewers can watch the low-frequency video content without being aware of this projector-camera communication cycle.

The process of IPT, including \textit{Modulation}, \textit{Demodulation}, and \textit{Pose Estimation}, is introduced in the following sections.

\setlength{\textfloatsep}{8pt plus 1.0pt minus 2.0pt}


\section{Modulation}

Modulation refers to the process that hides fiducial markers in a video in the form of high-frequency brightness change. This process utilizes the flicker-fusion property of Human Vision System (HVS). For example, if a screen alternates above the typical frequency of 40$\sim$50 Hz, human eyes cannot capture this fluctuation but perceive the average brightness instead \cite{simonson_flicker_1952}. Most laser projectors on the market have a vertical refresh rate of more than 48 Hz, which provides the hardware basis for the algorithm.

\subsection{Encoding Process}

\begin{figure}[t]
    \centerline{\includegraphics[scale=0.47]{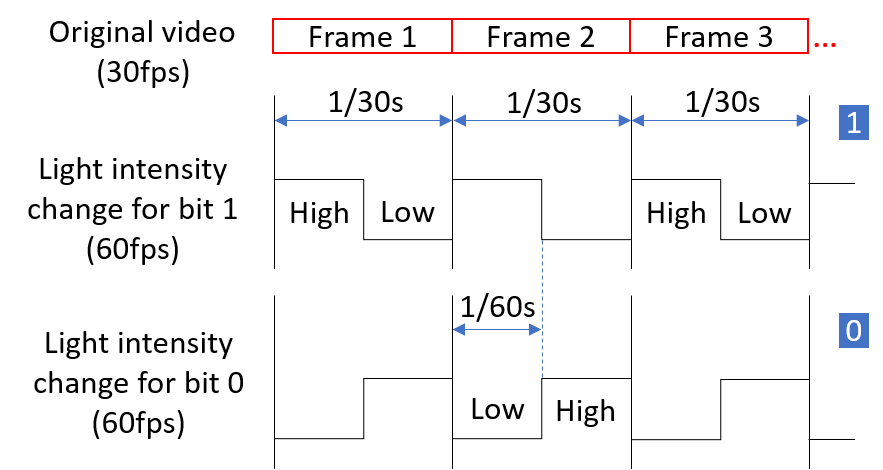}}
    \caption{The encoding process of binary pixels. Bit 0 and bit 1 are encoded as 60 FPS flashing in different phases. Since the flicker-fusion property, human eyes are difficult to perceive this high-speed lightness change.}
    \label{fig:encoding_process}
\end{figure}

Tags are hidden in the temporal dimension by converting them into time-variant high-frequency signals.
For a video at 30 FPS, the first step is to double each frame to get a new video at 60 FPS (or other frequency supported by the projector). Then the tags are embedded by adding or subtracting a light intensity $\Delta L$ on the pixel of the original frame at the corresponding position. The hidden pattern includes only 0 and 1, encoded similarly but in different phases, as shown in Fig. \ref{fig:encoding_process}.
This encoding scheme is called a Manchester-like coding scheme \cite{nguyen_high-rate_2016}, ensuring that the projector screen continuously blinks at high speed. The changing light intensity, $\Delta L$, is chosen according to the projector and the environment brightness. Under the guarantee of demodulating successfully, $\Delta L$ is required to be as small as possible to raise invisibility.

\setlength{\textfloatsep}{10pt plus 1.0pt minus 2.0pt}

The flicker fusion property of HVS is also called the low-pass filtering property \cite{wang_inframe_2015}, filtering temporal high-frequency signals. This property emphasizes the possibility of hiding patterns among high-speed flickering frames.


\subsection{Color Space Selection}

The next step is to determine the color space. Human eyes have different levels of sensitivity to brightness changes of different colors. In order to improve invisibility, the modulation process is required to ensure  \textit{perceptually uniform}, which means that the pixels with different colors have similar visual observations for the same brightness change. The previous research has shown that the brightness adjustment in CIELAB color space is considerably more uniform than HSL, YUV, and RGB spaces \cite{zhang_chromacode_2021}. CIELAB space expresses color as three channels: $L^*$, $a^*$, and $b^*$, where $L^*$ is perceptual lightness, and $a^*$ and $b^*$ indicate the colors of HVS. The brightness changes in the $L^*$ channel can be evenly distributed, making humans perceive a uniform lightness change. Therefore, we first convert the image from RGB to CIELAB space, then modify the lightness through $L^*$ channel, and finally convert the color space back. This conversion is also needed for the demodulation process.

\setlength{\textfloatsep}{8pt plus 1.0pt minus 2.0pt}
\begin{figure}[t]
    \centerline{\includegraphics[scale=0.40]{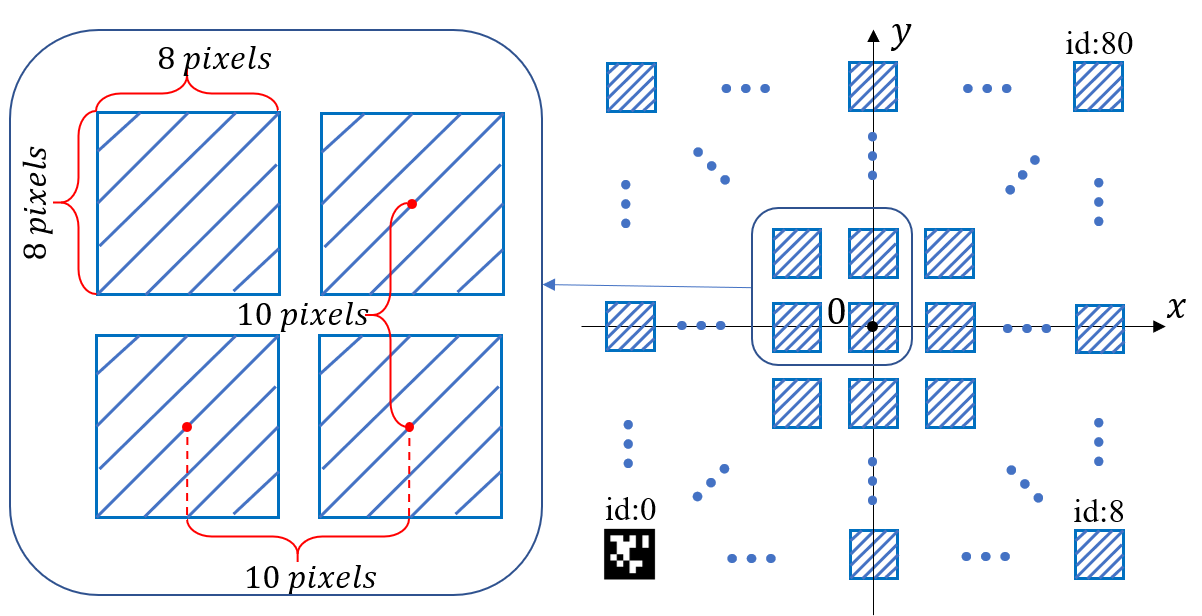}}
    \caption{A $9\times9$ binary tag map. The tag map is of the same size as the original video and designed for the localization task. After encoding, the tag map is coincided with the display image, making the coordinate systems of localization and visualization the same naturally.}
    \label{fig:tag_map}
\end{figure}

\begin{algorithm}[bt]
    \caption{Modulation}
    \label{alg:modulation}
    \begin{algorithmic}[1]
        \REQUIRE 1) $f_{i}$, the frequency of the input video; 2) $\bm{V}_i$, a video stream at $f_{i}$ frequency; 3) $M$, an AprilTag map; 4) $\Delta L$, the lightness change; 5) $w \times h$, the resolution of the output video; 6) $f_{o}$, the frequency of the output video
        \ENSURE $\bm{V}_o$, a modulated video
        \STATE $T_{mask} = M[white \leftarrow 1, black \leftarrow -1, others \leftarrow 0]$;
        \WHILE{Extract a frame $F$ from $\bm{V}_i$ successfully}
        \STATE $F \leftarrow$ resize($F$, $[w \times h]$);
        \STATE $F \leftarrow$ convertColor($F$, RGB2CIELAB);
        \STATE $N \leftarrow f_{o} / f_{i}$;
        \STATE $\textbf{F}=\{F_i\}, (i=1,2,...,N) \Leftarrow $ duplicate($F$) for $N$ times;
        \FOR{$F_i \in \textbf{F}$}
        \STATE $F_i[L] \leftarrow F_i[L] + \Delta L \cdot T_{mask}$, where $[L]$ indicates the lightness channel;
        \STATE $F_i[L>255] \leftarrow 255$ and $F_i[L<0] \leftarrow 0$;
        \STATE $\Delta L \leftarrow -1 \cdot \Delta L$;
        \STATE $F_i \leftarrow$ convertColor($F_i$, CIELAB2RGB);
        \ENDFOR
        \STATE $\bm{V}_o \Leftarrow \textbf{F}$.
        \ENDWHILE
    \end{algorithmic}
\end{algorithm}

\setlength{\dbltextfloatsep}{8pt plus 1.0pt minus 2.0pt}
\begin{figure*}[t]
    \centering
    \subfloat[Original image]{
        \includegraphics[width=.3\linewidth]{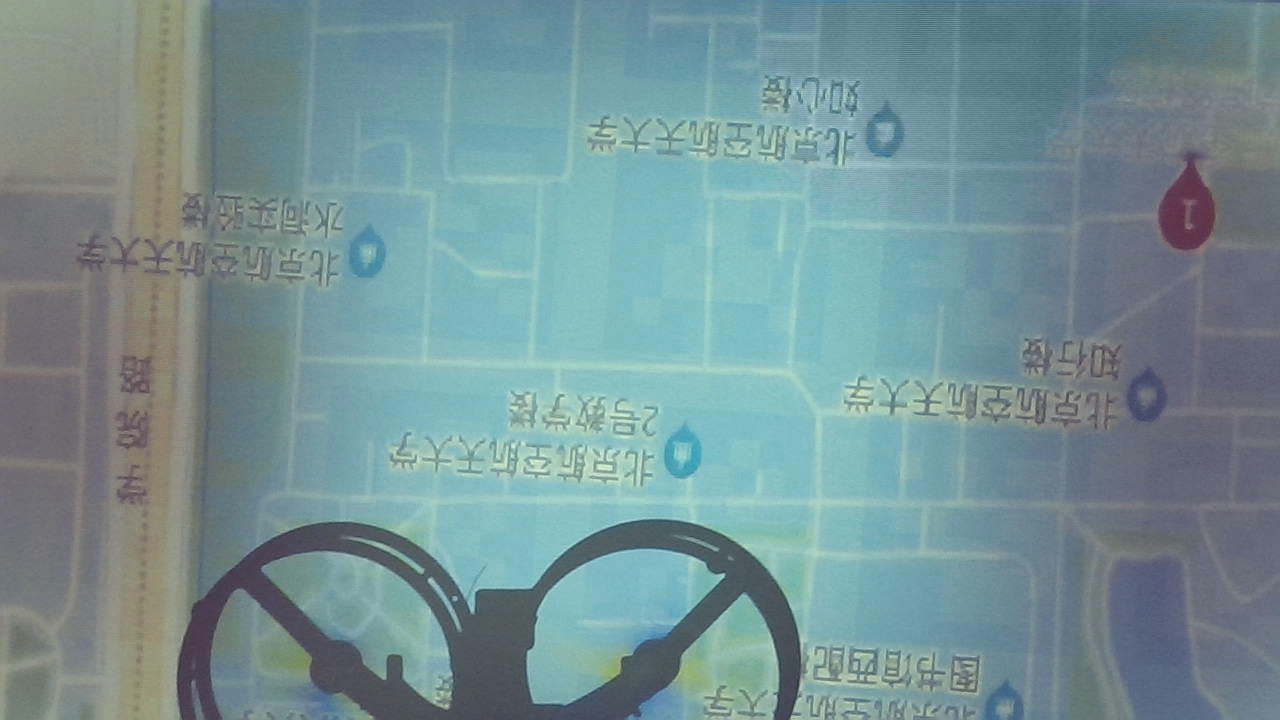}
        \label{fig:Original image}
    }\quad
    \subfloat[Subtract two successive frames]{
        \includegraphics[width=.3\linewidth]{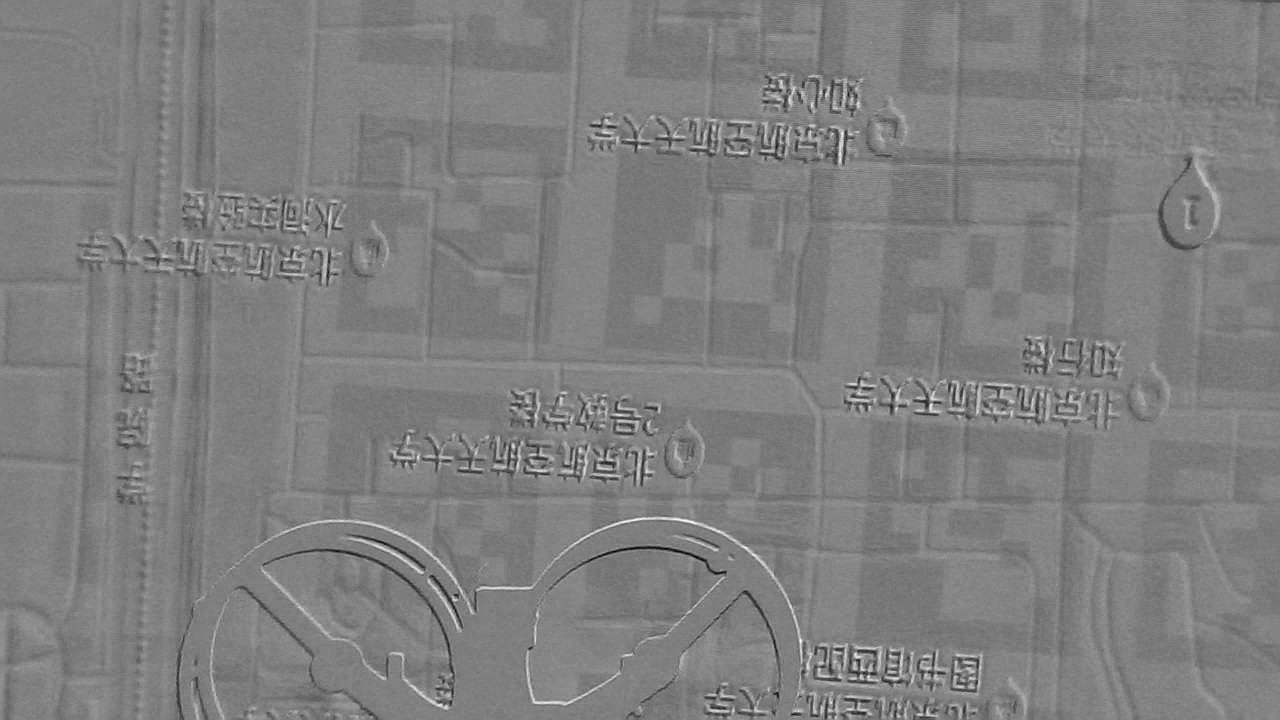}
        \label{fig:Without alignment}
    }\quad
    \subfloat[Image alignment]{
        \includegraphics[width=.3\linewidth]{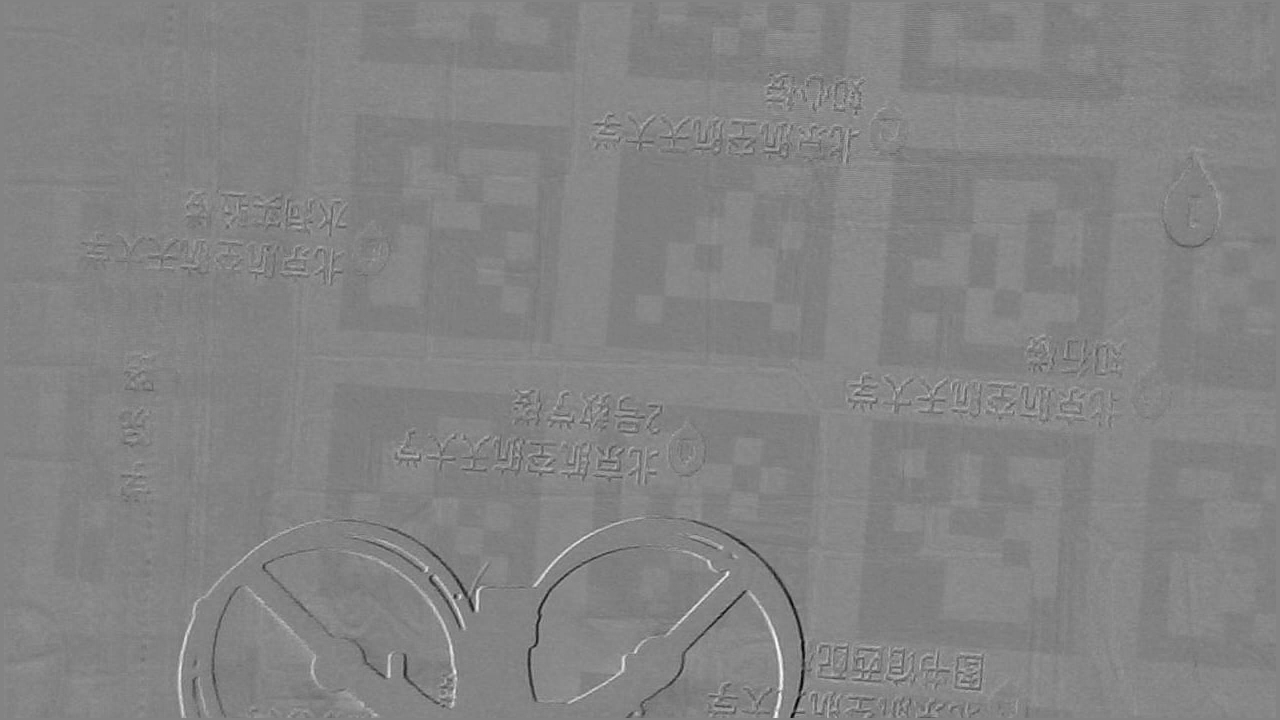}
        \label{fig:Shake elimination}
    }

    \centering
    \subfloat[Preprocessing]{
        \includegraphics[width=.3\linewidth]{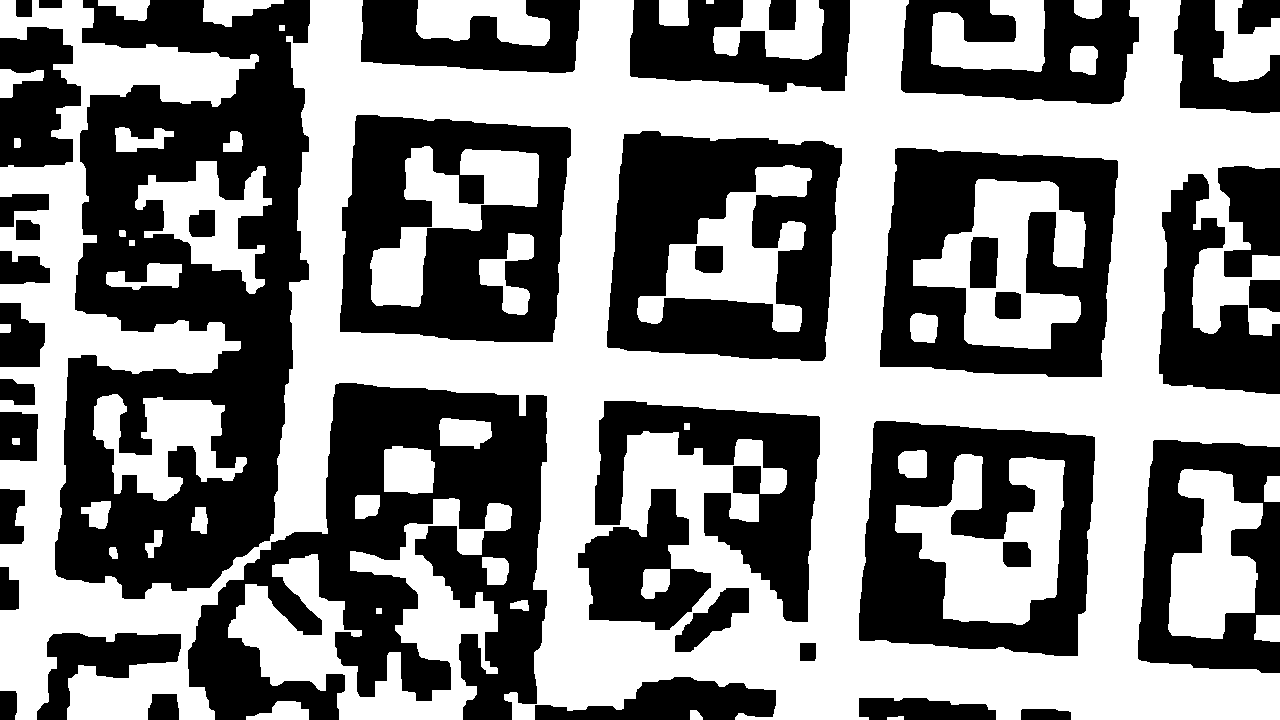}
        \label{fig:Preprocessing}
    }\quad
    \subfloat[Tag detection]{
        \includegraphics[width=.3\linewidth]{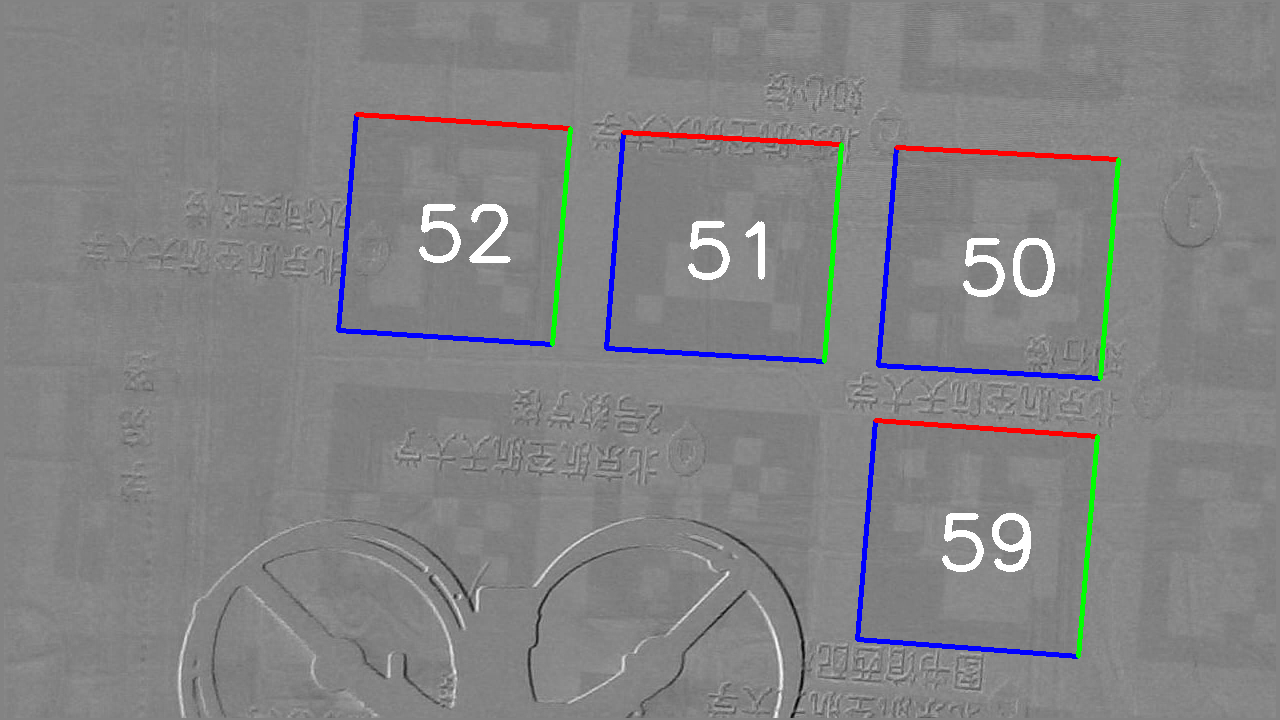}
        \label{fig:Tag detection}
    }

    \caption{Intermediate steps present the process of extracting fiducial tags from a raw image Fig. (a). First, the tags appear in Fig. (b) after the subtraction of two successive frames. Then, the disturbances such as text and lines are eliminated by image alignment, as shown in Fig. (c). Next, Fig. (d) displays the preprocessing result. Each tag forms a continuous boundary, ensuring the success of tag detection in Fig. (e). Although the shadow affects the tag detection in the left-down corner, other tags are still detected successfully. This shadow-tolerant character verifies the configuration of the tag map.}
    \label{fig:demodulation process}
\end{figure*}

\subsection{Tag Map Design}

Fig. \ref{fig:tag_map} shows how the position information is embedded in an image. The tag map uses the ENU coordinate system, of which the origin is the center of the map. The image is a black and white map covered with fiducial markers, and AprilTag is chosen here. Note that the tag density is the balance of projector's specifications (resolution and size of display area), flight conditions, and camera's properties (resolution and angle of view). If the tags are too large or too far apart, the tags captured by the camera are few. When the shadows of quadrotors occlude the tags, the demodulation process may fail. If the tags are too small or too close, the error of corner detection increases, reducing localization accuracy. Finally, the AprilTag map has $9\times9$ tags evenly distributed on a picture at a resolution of $1920\times 2160$. The tags are placed from left to right and from bottom to top.
Flight experiments have demonstrated that this configuration balances detection accuracy and robustness.

\vspace{6 pt}

The modulation algorithm is summarized in Algorithm \ref{alg:modulation}. After this process, the projector provides a suitable environment for the receivers to extract their locations.

\section{Demodulation}
The demodulation process first detects tags' IDs and their corner coordinates in the 2D image plane. Then tags' 3D world corner coordinates are calculated given IDs, the configuration of tag map, and the ratio from pixels to actual size. Note that this ratio must be measured before flights once the projector is fixed.

Many works have officially provided tools to detect tags. However, when the markers come from the subtraction of two successive frames, the quadrotor's movement causes many long and narrow lines at the edge of image patterns. These lines break the continuity of tags' outer black borders, causing detection failure. Therefore, we first propose a sample-based image alignment method and then a preprocessing step to prepare images, and finally, use an open-source tool AprilTag 3 for detection. The intermediate steps are shown in Fig. \ref{fig:demodulation process}.

\subsection{Image Alignment}

The frequency of the high-speed camera (120 FPS) is higher than the flicker frequency of the modulated video (60 FPS). Thus, after converting the color space from RGB to CIELAB, tags can be obtained by subtracting two successive frames in the lightness channel. However, the movement of quadrotors often causes sharp edges at the boundary of image texture or texts, as shown in Fig. \ref{fig:Without alignment}. These edges often cut off the tags. For AprilTag 3, an important step is to find the contour between white and black components \cite{krogius_flexible_2019}, so this discontinuity of the black borders greatly influences the detection. Therefore, the two frames need to be shifted a few pixels vertically and horizontally to eliminate as many edges as possible. The influence on localization accuracy can be ignored due to the very short-shifting distance of no more than 5 pixels.

In this work, a sample-based method is proposed to determine the shifting distance, achieving a trade-off between computation speed and alignment accuracy. For two images with a $w \times h$ resolution, sample $N$ columns from the two frames separately and denote them as $T^{pre}_i, (i=1,...,N)$ and $T^{now}_i, (i=1,...,N)$. The columns are equally spaced along the X-axis. The number of vertical shifting pixels, $m^*_y$, is expected to be the value when the difference between all $T^{pre}_i$ and $T^{now}_i$ achieves the least:
\begin{equation}
    \setlength\abovedisplayskip{3pt}
    \setlength\belowdisplayskip{4pt}
    \begin{aligned}
        m^*_y ~~~                         & =       \\[-2pt]
        \underset{|m_y| \leq b}{\arg\min} & \left\{
        \begin{aligned}
            \sum_{i=1}^{N} \| T^{pre}_i(0:h-m_y) - T^{now}_i(m_y:h) \|_1         & \\[-8pt]
            (m_y \geq 0)                                                         & \\[-8pt]
            \sum_{i=1}^{N} \| T^{pre}_i(-m_y\!:\!h)\!-\!T^{now}_i(0:h+m_y)  \|_1 & \\[-8pt]
            (m_y < 0)                                                            &
        \end{aligned}
        \right.
    \end{aligned},
\end{equation}
where $b$ is the maximum shifting distance set by the user, and $(0\!:\!h\!-\!m_y)$ denotes the elements from $0$ to $h\!-\!m_y\!-\!1$. Considering the balance of computational speed and effect, $N=3$ is chosen. The horizontal shifting value can be calculated through a similar method.

After computing the shifting value, the two shifted frames are subtracted to get a new image. Unfortunately, shifting also causes straight lines at image borders, which must be eliminated. The final aligned image is shown in Fig. \ref{fig:Shake elimination}. Comparing with the no-shifting Fig. \ref{fig:Without alignment}, the edges in Fig. \ref{fig:Shake elimination} are reduced considerably, proving the effect of the algorithm.

\subsection{Preprocessing}

Preprocessing includes normalization, filtering, binarization, and morphology operations. Among these steps, the most critical parameter is an appropriate threshold for binarization. Since the edges caused by motion always appear in pairs of black and white, and the tags have approximately equal areas of black and white pixels, the median of the image is chosen as the threshold. After thresholding, 'OPEN' operation is first applied to remove the noise further, and 'CLOSE' operation is then used to connect the borders of the tags, shown as Fig. \ref{fig:Preprocessing}.

\subsection{Tag Detection}

The article \cite{kallwies_determining_2020} has extensively compared four widely used tag detection algorithms (AprilTag 3, AprilTags C++, ArUco, and ArUco OpenCV) on localization accuracy and processing time. From the article, AprilTag 3 is the best choice in most cases. Therefore, AprilTag 3 is selected as the tag detection tool, and the \textit{tag36h11} set is chosen as tag family. The detecting result is shown in Fig. \ref{fig:Tag detection}.


\vspace{6 pt}

The demodulation algorithm is summarized in Algorithm \ref{alg:demodulation}. The 2D coordinates and the 3D coordinates of tag corners are used in the next section.

\begin{algorithm}[bt]
    \caption{Demodulation}
    \label{alg:demodulation}
    \begin{algorithmic}[1]
        \REQUIRE 1) $F^{now}$, an RGB frame from camera; 2) $F_L^{pre}$, the lightness channel of previous frame; 3) $C_{map}$, the configuration of the AprilTag map; 4) $r$, the ratio from pixels to real size.
        \ENSURE $\bm{R}$, a result list that contains id, corner points, hamming distance and world coordinates for each tag.
        \WHILE{receive a new $F_L^{now}$}
        \STATE $F^{now} \leftarrow$ convertColor($F^{now}$, RGB2CIELAB);
        \STATE $F_L^{now} \leftarrow$ $F^{now}[L]$, where $[L]$ indicates the lightness channel;
        \STATE $F_L^{now'} \leftarrow$ align($F_L^{now}$, $F_L^{pre}$);
        \STATE $F_b \leftarrow$ normalization($F_L^{now'}-F_L^{pre}$);
        \STATE $F_L^{pre} \leftarrow  F_L^{now}$, for the next loop;
        \STATE $F_b \leftarrow$ meanFilter($F_b$);
        \STATE $F_b \leftarrow$ threshold($F_b$, median($F_b$));
        \STATE $F_b \leftarrow$ morphologyEx($F_b$, OPENING);
        \STATE $F_c \leftarrow$ morphologyEx($F_b$, CLOSING);

        \STATE $\{id, p_{i}, ham\} \leftarrow$ detectTags($F_c$);
        \STATE $\{p_{w}\} \leftarrow$ lookupMap($\{id\}$, $C_{map}$, $r$);
        \STATE Return $\bm{R} \leftarrow \{id, p_{i}, ham, p_{w}\}$.
        \ENDWHILE
    \end{algorithmic}
\end{algorithm}

\section{Pose Estimation}

The pose estimation process based on fiducial markers is called a \textit{Solve PnP} problem. It means to solve the position and orientation of a camera giving the 3D coordinates of n points in space and their corresponding 2D coordinates in the photo plane. The core of this problem is to solve (2):
\begin{equation}
    \setlength\abovedisplayskip{2pt} 
    \setlength\belowdisplayskip{0pt}
    z_{c}\left[\begin{array}{l}
            u \\
            v \\
            1
        \end{array}\right]=\left[\begin{array}{lll}
            f_{x} & 0     & u_{0} \\
            0     & f_{y} & v_{0} \\
            0     & 0     & 1
        \end{array}\right]\left[\begin{array}{ll}
            R \ |\  \vec{T}
        \end{array}\right]\left[\begin{array}{c}
            x_{w} \\
            y_{w} \\
            z_{w} \\
            1
        \end{array}\right],
\end{equation}
that is
\begin{equation}
    \setlength\abovedisplayskip{0pt}
    \setlength\belowdisplayskip{0pt}
    {z_c} \cdot {p_i}= {M_1} \cdot {M_2} \cdot {p_w},
\end{equation}
where $z_c$ is the scale factor, $[u, v]^{T}$ and $p_i$ are the 2D photo coordinates, $f_x$ and $f_y$ are the scaled focal lengths, $[u_0, v_0]^{T}$ is the principal point, $M_1$ is the matrix of intrinsic camera parameters, $R$ and ${\vec T}$ are the rotation matrix and position vector from the world frame to the camera frame, $M_2$ is the 3D pose matrix, as well as $[x_w, y_w, z_w]^{T}$ and $p_w$ refer to the world 3D coordinates.

In (3), coordinates $p_i$ and $p_w$ are calculated through the demodulation algorithm in the last section, and the camera matrix $M_1$ can be obtained by camera calibration.
Considering the performance and stability, we use IPPE algorithm \cite{collins_infinitesimal_2014} to compute the pose matrix $M_2$.
Note that the results from {\tt\small solvePnP()} function in OpenCV must be inversely transformed using (4):
\begin{equation}
    \setlength\abovedisplayskip{3pt}
    \setlength\belowdisplayskip{3pt}
    \left[\begin{array}{ll}
            ^w_cR \ |\  ^w\vec{T}
        \end{array}\right]=\left[\begin{array}{ll}
            R^T \ |\  -R^T \cdot \vec{T}
        \end{array}\right],
\end{equation}
where $^w_cR$ and $^w\vec{T}$ are the rotation and translation from the camera frame to the world frame, respectively.

\section{Experiments}

\setlength{\textfloatsep}{8pt plus 1.0pt minus 2.0pt}
\begin{figure}[t]
    \centerline{\includegraphics[scale=0.42]{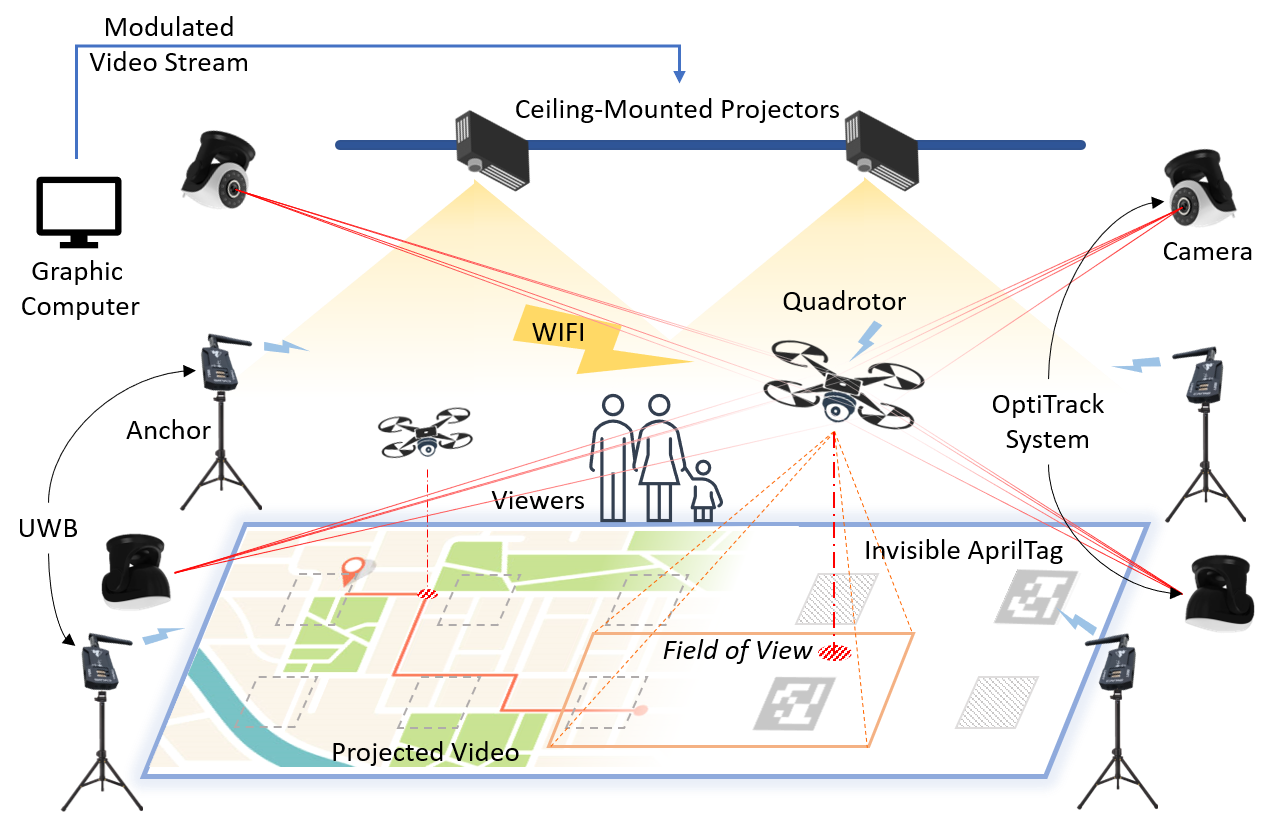}}
    \caption{IPT AR flight system, adding UWB and OptiTrack for comparison. The UWB system includes one tag on the quadrotor and four base anchors. OptiTrack uses a dozen of cameras to provide submillimeter-level position.}
    \label{fig:experiment_system}
\end{figure}

\subsection{Hardware System}

We make an experiment system to test the performance of IPT. The system is illustrated in Fig. \ref{fig:experiment_system}, including hardware for both sender and receiver parts. The sender part uses two NEC CR3400HL laser projectors to form a $1920\!\times\!2160$ screen, of which the size is $2.17 {\rm m} \times 2.47 {\rm m}$. An NVIDIA 1080 TI graphics card and NVIDIA \textit{Surround} function are utilized to output a $1920\!\times\!2160$ 60-Hz video stream. The receiver part is a 270-mm-wheelbase quadrotor with a high-speed camera. To eliminate \textit{rolling shutter effect}, which means two successive frames are mixed with each other, a 120-FPS color global shutter camera is selected. The onboard computer is a Raspberry Pi 4B with a Broadcom BCM2711 quad-core CPU running at 1.5 GHz.

The system also involves a Nooploop UWB system (25 fps) and an OptiTrack system (120 fps) for performance comparison. We carry out several indoor flights on this system to test the performance of IPT.

\subsection{Experimental Setup}

The experiment consists of two stages. The first stage aims to test accuracy. For convenience, all pictures shot by the camera are stored and then post-processed by a personal computer. We choose a 15-second video as the test unit, including 1800 images in total. The second stage aims to test processing speed, and hence the pictures are processed onboard for real-time computation.

To simulate the path planning mission, we choose the path searching process on Amap as the video content and record it for 30 seconds in advance. This screen recording is looped as the sender video. In the experiment, the quadrotor flies in a rectangle by a pilot at a height of about 0.8 m.

\setlength{\textfloatsep}{10pt plus 1.0pt minus 2.0pt}
\begin{figure}[t]
    \centering
    \subfloat[Position comparison]{
        \includegraphics[width=0.485\textwidth, trim={0 2mm 0 1.5mm}, clip]{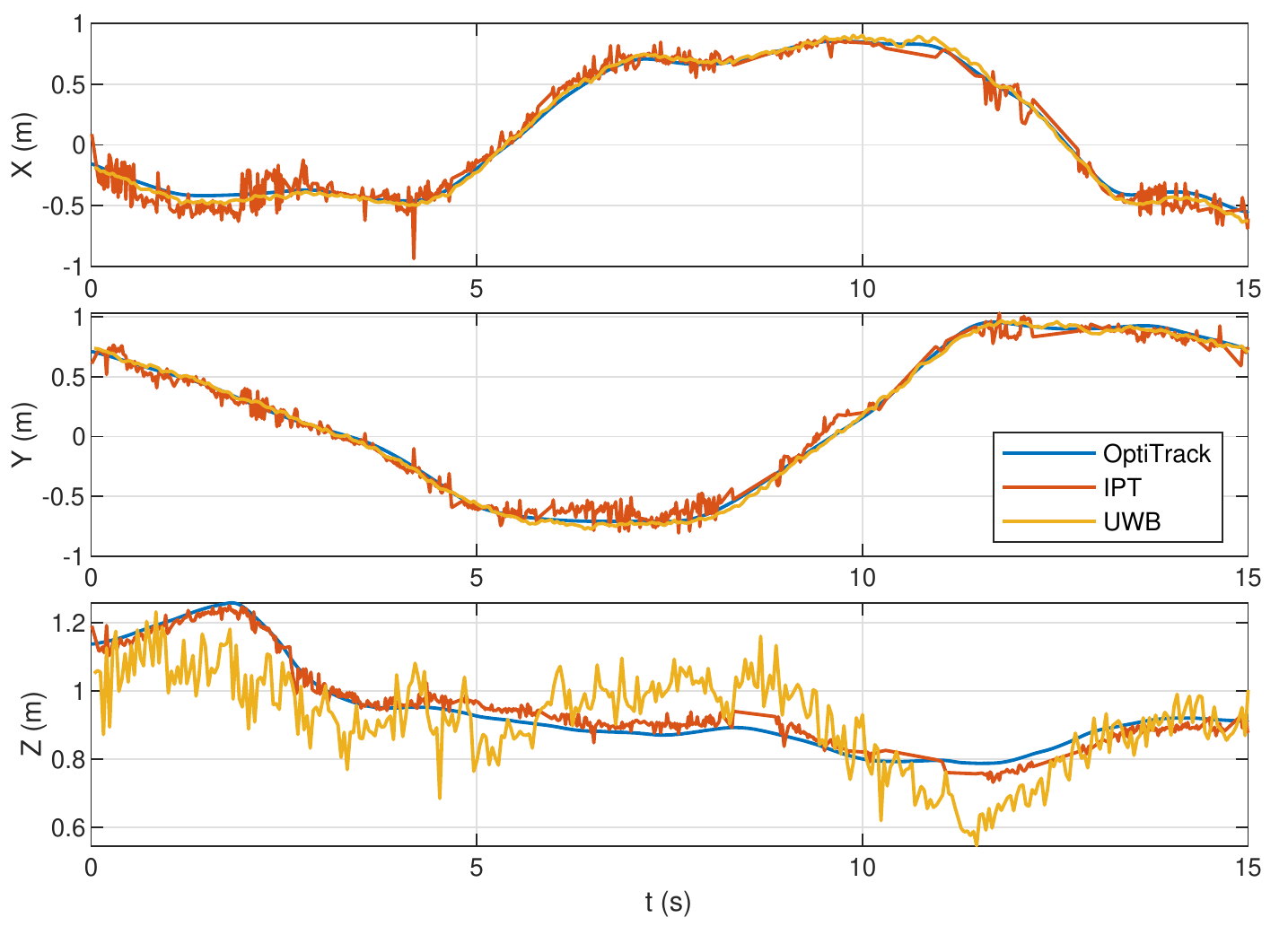}
        \label{fig:posi_cpr}
    }
    \\
    \subfloat[Orientation comparison]{
        \includegraphics[width=0.485\textwidth, trim={0 2mm 0 1.5mm}, clip]{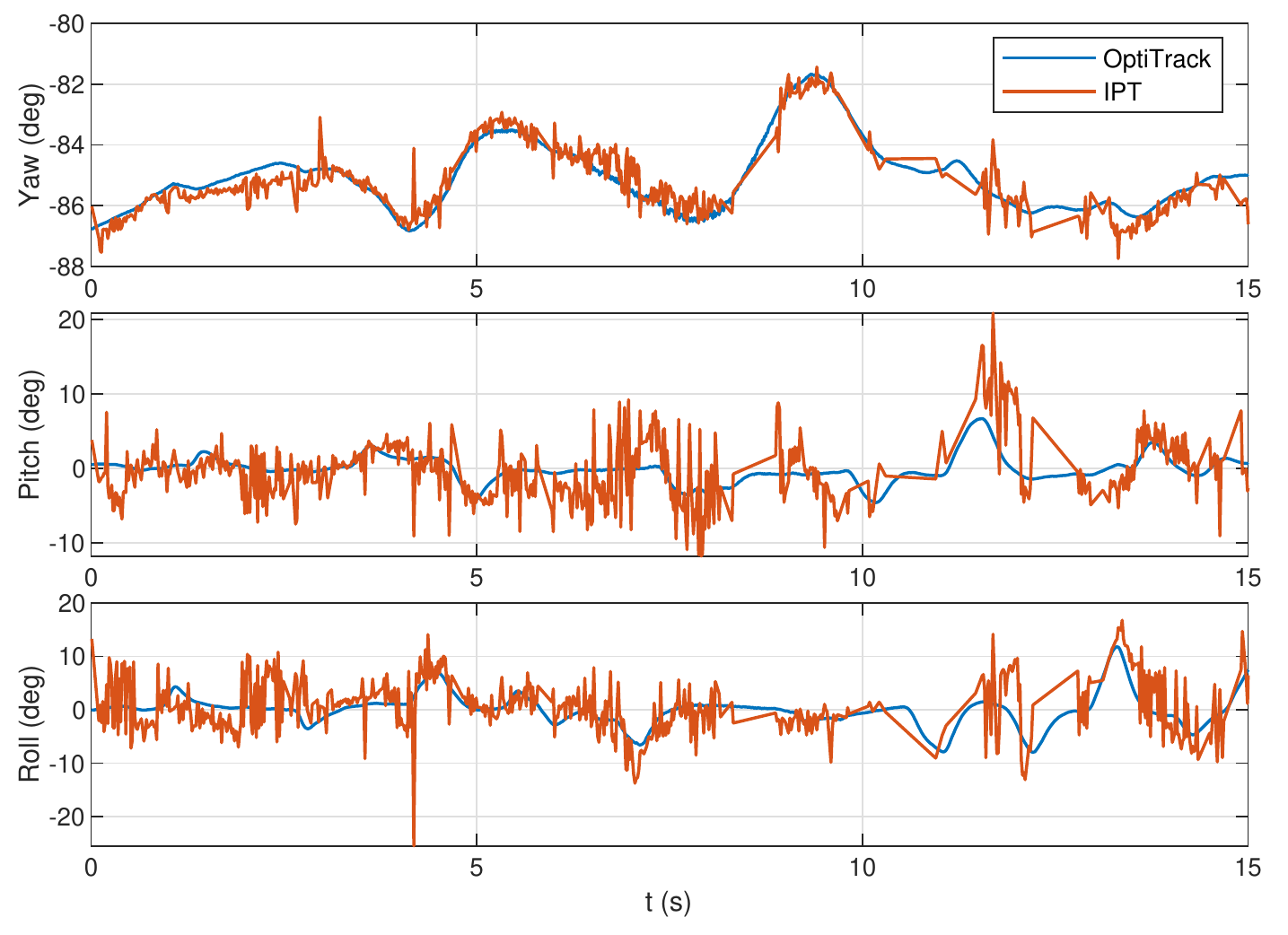}
        \label{fig:ori_cpr}
    }
    \\
    \subfloat[Positioning error of IPT and velocity varying over time]{
        \includegraphics[width=0.48\textwidth, trim={0 1mm 0 0}, clip]{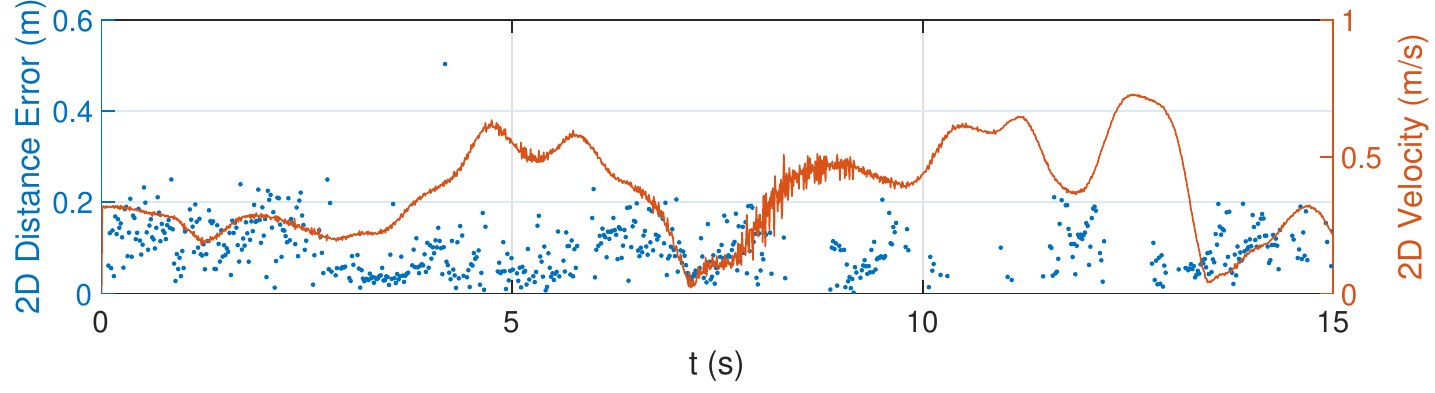}
        \label{fig:error_velocity}
    }

    \caption{The comparison of OptiTrack, IPT, and UWB data for a 15-second indoor flight. In Fig. (a), IPT achieves better accuracy in the Z direction than UWB but more oscillations in the X and Y directions. In Fig. (b), UWB cannot measure orientation, but IPT obtains acceptable yaw angle and oscillating curves in pitch and roll. Finally, Fig. (c) illustrates the relationship between horizontal positioning error and velocity change.}
    \label{fig:localization comparison}
\end{figure}

\subsection{Results}

\subsubsection{Visual Effect}

We first invite three colleagues to evaluate the visual effect of IPT. The invisibility of markers is closely related to the changing light intensity $\Delta L$. When $\Delta L=3$, they hardly notice the markers. When $\Delta L=4$, they observe that something square is slightly blinking but cannot distinguish what it is. When  $\Delta L=5$, the tag map is obvious. We choose $\Delta L=4$ for the following tests considering the balance of invisibility and demodulation effect.

\begin{table}[t]
    \setlength{\abovecaptionskip}{0pt}
    \setlength{\belowcaptionskip}{-2pt}
    \centering
    \caption{Positional error and rotational error (\textbf{metrics}: mae / std).}
    \begin{tabular}{ l|c|c|c }
        \toprule  
        ~   & \textbf{X Error [m]}     & \textbf{Y Error [m]}       & \textbf{Z Error [m]}      \\
        \midrule  
        UWB & \textbf{0.0305 / 0.0354} & \textbf{0.0232 / 0.0251}   & 0.0943 / 0.1128           \\
        IPT & 0.0690 / 0.0886          & 0.0535 / 0.0674            & \textbf{0.0230 / 0.0263}  \\
        \bottomrule 
        \toprule
        ~   & \textbf{Yaw Error [deg]} & \textbf{Pitch Error [deg]} & \textbf{Roll Error [deg]} \\
        \midrule
        UWB & —                        & —                          & —                         \\
        IPT & \textbf{0.352 / 0.460}   & \textbf{2.659 / 3.657}     & \textbf{3.342 / 4.305}    \\
        \bottomrule
    \end{tabular}
    \label{tab:error results}
\end{table}

\subsubsection{Localization Accuracy}

During the experiment, 625 data are sampled successfully for the IPT method. In order to display the raw effect, data are not filtered. Aligning the data from OptiTrack, IPT, and UWB in the dimension of time and space, we draw the comparison result in Fig. \ref{fig:localization comparison}.
OptiTrack is set as the ground truth since it can offer submillimeter-level position and 0.01-degree-level orientation.
To quantitatively analyze, Mean Absolute Error (MAE) and Standard Deviation (STD) of UWB and IPT are calculated in Table \ref{tab:error results}.
From the table and Fig. \ref{fig:posi_cpr}, IPT attains better position data in the Z direction than UWB but much more oscillated in the X and Y directions. For orientation, UWB cannot measure this state but IPT obtains acceptable yaw angle and volatile pitch angle and roll angle.
Note that the oscillations for horizontal positions mainly comes from the last step of \textit{Pose Estimation}, because the severe oscillations in pitch and roll bring fluctuations in inversely coordinates transformation.


The relation of horizontal positional error and velocity is shown in Fig. \ref{fig:error_velocity}, of which each blue point indicates a successful localization. Observing the time interval of 4-7s, the algorithm collects many samples even at a relatively high velocity, which verifies the image alignment algorithm.




\begin{table}[t]
    \setlength{\abovecaptionskip}{0pt}
    \setlength{\belowcaptionskip}{-2pt}
    \centering
    \caption{The comparison of three localization methods.}
    \begin{tabular}{ l|c|c|c }
        \toprule
                                    & \textbf{OptiTrack} & \textbf{UWB} & \textbf{IPT}        \\
        \midrule
        \textbf{Price\footnotemark} & CNY 190,000        & CNY 35,000   & \textbf{CNY 30,000} \\
        \textbf{Position Accuracy}  & sub-mm level       & cm level     & cm level            \\
        \textbf{Orientation Info}   & YES                & NO           & \textbf{YES}        \\
        \textbf{Anti-Interference}  & YES                & NO           & \textbf{YES}        \\
        \textbf{No Calibration}     & NO                 & NO           & \textbf{YES}        \\
        \bottomrule
    \end{tabular}
    \label{tab:comparison para}
\end{table}

\subsubsection{Computational Speed}

A C++ implementation of IPT achieves real-time processing at 10 FPS on average for $640 \times 360$ resolution. This frequency underlines the possibility for flight control of quadrotors.

\vspace{6 pt}

The comparison of OptiTrack, UWB, and IPT methods for an AR robotics platform is listed in Table \ref{tab:comparison para}. IPT achieves centimeter-level position and orientation measurements with advantages on price and efficiency, making it a competitive choice for localization in indoor AR environments.

\footnotetext{This price consists of devices for both localization and visualization but no personal computers and robots.}


\section{Conclusion}

We proposed IPT, an indoor localization method based on human-invisible projected fiducial tags. This method utilized the visual characteristics of human eyes to hide markers. To test the performance, we designed an AR quadrotor flight platform, of which the structure consists of the sender part and receiver part. Indoor flight experiments were conducted to evaluate accuracy and speed of IPT. The result presented centimeter-level accuracy and a frequency of ten FPS. The economical and quick-start features made IPT an appropriate solution for AR robotics platforms.
In future work, we plan to integrate IPT with IMU to obtain more accurate and high-frequency pose information. We will also explore the effect of IPT on different videos and the relationship between movement and performance.

\section*{Acknowledgments}

We sincerely thank anonymous reviewers for their critical reading and helpful suggestions.
This work is supported by the Science and Technology Innovation 2030-Key Project of ``New Generation Artiﬁcial Intelligence" under Grant 2018AAA0102305, the Zhejiang Provincial Natural Science Foundation of China under Grant No.LGG22F030025, and the National Natural Science Foundation of China under Grants 61803014, 61922008, and 61873011.

\clearpage
\bibliographystyle{IEEEtranBST/IEEEtran}
\bibliography{IEEEtranBST/IEEEabrv, IEEEtranBST/references}

\end{document}